\pdfoutput=1
\documentclass[11pt]{article}

\usepackage[final]{acl}

\usepackage{times}
\usepackage{latexsym}

\usepackage[T1]{fontenc}
\usepackage[utf8]{inputenc}

\usepackage{microtype}

\usepackage{inconsolata}

\usepackage{graphicx}
\usepackage{algorithm, algpseudocode}
\usepackage{xcolor}
\usepackage{mathtools}
\usepackage{multirow}
\usepackage{makecell}
\usepackage{bbm} 
\usepackage{amsmath}
\usepackage{amssymb}
\usepackage{xcolor}
\usepackage{tabularx}
\usepackage{array}
\usepackage{kotex}
\usepackage{colortbl}

\usepackage{tikz,times}
\usetikzlibrary{positioning}
\usetikzlibrary{snakes}
\usetikzlibrary{arrows,intersections}
\usetikzlibrary{mindmap,backgrounds}
\tikzstyle{block} = [draw, rectangle,
minimum height=3em, minimum width=5em, align=left]
\tikzstyle{pinstyle} = [pin edge={to-,thin,black}]
\usepackage{mdframed} % Include this in your preamble

\def\D{{\cal D}}

\def\P{{\cal P}}

\title{SGuard-v1: Safety Guardrail for Large Language Models}

\author{
 \textbf{JoonHo Lee}\textsuperscript{*},
 \textbf{HyeonMin Cho}\textsuperscript{*},
 \textbf{Jaewoong Yun}\textsuperscript{*},
 \textbf{Hyunjae Lee}\textsuperscript{*},
 \textbf{JunKyu Lee}\textsuperscript{*}
and
 \textbf{Juree Seok}\textsuperscript{*}
 \\
\\
 \textsuperscript{}Samsung SDS Technology Research
\\
 \small{
   \textbf{Correspondence:} \href{mailto:joonholee@samsung.com}{joonholee@samsung.com}
 }
}

\begin{document}
\maketitle

\begingroup
\renewcommand\thefootnote{\hspace*{-0.8em}*}
\footnotetext{Equal contribution}
\endgroup

\begin{abstract}
We present SGuard-v1, a lightweight safety guardrail for Large Language Models (LLMs), which comprises two specialized models to detect harmful content and screen adversarial prompts in human–AI conversational settings.
The first component, ContentFilter, is trained to identify safety risks in LLM prompts and responses in accordance with the MLCommons hazard taxonomy, a comprehensive framework for trust and safety assessment of AI. 
The second component, JailbreakFilter, is trained with a carefully designed curriculum over integrated datasets and findings from prior work on adversarial prompting, covering 60 major attack types while mitigating false-unsafe classification.
SGuard-v1 is built on the 2B-parameter Granite-3.3-2B-Instruct model that supports 12 languages. 
We curate approximately 1.4 million training instances from both collected and synthesized data and perform instruction tuning on the base model, distributing the curated data across the two component according to their designated functions. 
Through extensive evaluation on public and proprietary safety benchmarks, SGuard-v1 achieves state-of-the-art safety performance while remaining lightweight, thereby reducing deployment overhead.
SGuard-v1 also improves interpretability for downstream use by providing multi-class safety predictions and their binary confidence scores. 
We release SGuard-v1 \href{https://huggingface.co/collections/SamsungSDS-Research/sguard}
{here} under the Apache-2.0 License to enable further research and practical deployment in AI safety.

\end{abstract}
\noindent

\begin{quote}
\small
\textbf{Content Warning.}
This paper contains verbatim examples of harmful language used for research.
\end{quote}

\section{Introduction}
\label{sec:intro}

Large Language Models (LLMs) have demonstrated innovative performance across a wide range of tasks, from natural language understanding to creative content generation, and transformed both industry and research practice.
However, deploying LLMs in real applications introduces safety risks such as physical, non-physical, and contextual hazards that demand rigorous evaluation and careful deployment strategies \cite{ailuminate}. 
Safety alignment within the model itself is the primary mitigation strategy, but it remains imperfect and leaves several vulnerabilities.
In particular, jailbreak attacks—where an adversarially crafting prompts that bypass LLMs' safety alignment and induce illegal, biased, or unethical outputs—have emerged as a serious threat during human-AI interactions.

\begin{figure}[!t]
  \centering
  \includegraphics[width=1.0\columnwidth]{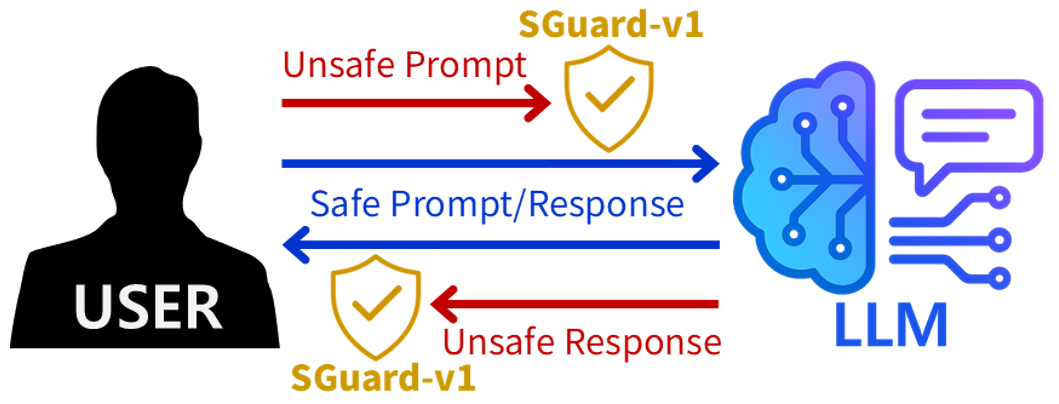}
% \vspace{-1.4ex}
\caption{
(Best viewed in color)
The schematic illustration of SGuard-v1: Harmful and adversarial prompts are screened by ContentFilter and JailbreakFilter while unsafe responses generated by LLMs are filtered by ContentFilter.
{\footnotesize $^*$The LLM image is generated by GPT-5.}
}
\label{fig:overview}
\end{figure}

The growing prevalence and sophistication of these threats underscores the need for robust defense mechanisms. 
As a result, most generative AI systems are now encouraged to incorporate safety guardrails that operate at the input and output level \cite{llamaguard, llamaguard3vision, aws_bedrock, openai_mod}. 
These guardrails form an external safety layer that enforces policy without modifying the model’s internal weights, and are designed to block harmful contents and behaviors including adversarial jailbreaking attempts as well as the generated responses that are safety-critical or policy-violating as illustrated in Figure \ref{fig:overview}.
%------------------------------------------------------------------------
%------------------------------------------------------------------------
\begin{table*}[!t]
\footnotesize
\centering
\def\arraystretch{1.15}
\caption{Safety risk categories in our ContentFilter model}
 \vspace{-1ex}
\resizebox{0.95\textwidth}{!}{%
\begin{tabularx}{1.0\textwidth}{>{\raggedright\arraybackslash}p{0.20\linewidth} | >{\raggedright\arraybackslash}X}
\Xhline{2\arrayrulewidth}
Category  & Definition \\
\hline
Violence and Hate 
& Content that promotes or praises physical or psychological harm to others, incites violence, or contains hateful, discriminatory, or harassing expressions targeting an individual or group \\
\hline
Illegal and Criminal Activities  
& Content that encourages or instructs others to engage in illegal behavior, supports or plans unlawful activity, or provides guidance intended to facilitate criminal conduct \\
\hline
Sexual Content and Exploitation 
& Content that includes explicit sexual descriptions or depicts sexually inappropriate material involving minors, including sexualization of minors \\
\hline
Privacy and Sensitive Information Misuse  
& Content that discloses or seeks to disclose sensitive personally identifiable information without consent that enables doxxing or unauthorized account access, leaks proprietary or confidential data, or provides unqualified guidance on health, legal, or financial matters \\
\hline
Manipulation and Societal Harm  
& Content that spreads false or misleading narratives (e.g., conspiracy theories, disinformation), promotes extremist propaganda or political manipulation, or attempts to erode public trust through deception or targeted influence \\
\Xhline{2\arrayrulewidth}
\end{tabularx}
}%
\vspace{-1ex}
\label{tab:taxonomy}
\end{table*}
%------------------------------------------------------------------------

Our goal is to develop a high-performance bilingual safety guardrail that can be deployed in real time across diverse LLM platforms with low memory footprint and latency, while maintaining high accuracy in harmful content detection for human-LLM conversation. 
To this end, we design filter models on top of Granite-3.3-2B-Instruct \cite{granite}, which is comparatively lighter than the LLMs used in existing approaches, and we accordingly invest in more rigorous refinement of English and Korean data and the training pipeline.

The proposed safety guardrail SGuard-v1 includes ContentFilter as a primary component. 
ContentFilter is trained to assess the safety of both user inputs and model outputs with high accuracy, and to immediately trigger blocking or mitigating actions based on that assessment. 
In addition, we provide JailbreakFilter, which is specialized for defending against sophisticated evasion and jailbreak attempts that intend to elicit unsafe or policy-violating responses.
Combining these two components enables LLM-based services to maintain robust safety alignment that provides a foundation for reliable AI deployment.
Extensive evaluation across public and proprietary safety benchmarks show that SGuard-v1 achieves state-of-the-art performance and while maintaining lightweight footprint.
We release SGuard-v1 under the Apache-2.0 License at \href{https://huggingface.co/collections/SamsungSDS-Research/sguard}
{this repository} to enable both academic use and real-world deployment for AI safety by AI researchers and practitioners.
% \footnote{https://huggingface.co/collections/SamsungSDS-Research/sguard}.
Our key contributions are summarized as follows:
\begin{itemize}
\item We introduce SGuard-v1, a lightweight dual-guardrail solution that consists of ContentFilter for input/output moderation and JailbreakFilter for adversarial defense.
\item We present a bilingual training pipeline including data curation and curriculum design that lead to strong performance in English and Korean safety benchmarks.
\item We release SGuard-v1 under the Apache-2.0 License to advance research on AI safety and to support AI practitioners in deploying safer LLM-based systems.
\end{itemize}
%------------------------------------------------------------------------

\section{Safety Risk Taxonomy}
\label{sec:taxonomy}
As LLMs become part of safety-critical systems, many government and industry organizations such as \citet{eu_ai_act, nist_ai_600_1, ailuminate} provide policies and guidelines for classifying safety risks. 
Among these efforts, \citet{ailuminate} introduces an open, community-driven benchmark suite developed in collaboration with more than 50 participating organizations across industry, academia, and civil society. The suite evaluates the safety and reliability of general-purpose AI systems across 12 standardized hazard categories: Violent Crimes, Sex-Related Crimes, Child Sexual Exploitation, Suicide \& Self-Harm, Indiscriminate Weapons, Intellectual Property, Defamation, Non-Violent Crimes, Hate, Privacy, Specialized Advice, and Sexual Content.

While preserving compatibility with the taxonomy of \citet{ailuminate}, we consolidate the original twelve categories into five broader groups as shown in Table \ref{tab:taxonomy} and train our ContentFilter model using these safety risk categories to improve training and inference efficiency of the ContentFilter.
Our JailbreakFilter model is trained on over one million examples spanning 60 major attack types and learns to produce safe-unsafe binary decisions with associated confidence scores.

\section{Building SGuard-v1}
\label{sec:building}
\begin{center}
\fcolorbox{red!50}{red!10}{%
  \parbox{0.95\linewidth}{%
    \small
    \textbf{Content Warning.}
    This section, particularly Figures~\ref{fig:cht_example} and~\ref{fig:bhcb_example}, includes verbatim examples of toxic language.
  }%
}
\end{center}

\noindent
In this section, we introduce the training data curation and training pipeline of the ContentFilter and JailbreakFilter models in SGuard-v1.
We note that the models are explicitly trained to support English and Korean and their performance is evaluated in these two languages. 
While we do not claim reliable coverage across all 12 languages supported by the base model, we observed that the models retain a non-trivial level of capability in several of those additional languages.

\subsection{SGuard-ContentFilter-2B-v1}
\label{sec:ContentFilter}
\paragraph{Data Preparation}
We use WildGuardMix \cite{han2024wildguard} and Aegis \cite{aegis20} as our seed datasets, which have 86,759 and 30,007 English examples with safe\footnote{We use the terms "safe" and "benign" interchangeably.}-unsafe binary annotations, respectively. 
Since we focus on bilingual capability of our models, we translate the seed datasets into Korean through our CHT (Contextual Harm Translation) method as follows:
For a given English text $x_\text{source}$, we obtain $x^{'}=T(x_\text{source})$ by translating $x_\text{source}$ using a translator $T$ ({\it e.g.} DeepL\footnote{https://www.deepl.com/}). 
As LLM-based translators often reject to respond with translated harmful contents because of their own safety alignment, the translated $x^{'}$ often contains different (even opposite) meaning. 
To tackle this problem, we apply context engineering that can bypass the safety alignment of translators by employing a refiner function, $R(x_\text{source}, x^{'})$.
To obtain the final translation candidate $x_\text{target}$, the prompt for $R$ instructs to compare $x_\text{source}$ and $x^{'}$, and if $x^{'}$ is semantically different from $x_\text{source}$, the prompt encourages $R$ to return a revised $x^{'}$ to be semantically aligned with $x_\text{source}$. 
In our CHT approach, we use a recent 70B-parameter instruction-tuned LLM as $R$.
Figure \ref{fig:cht_example} shows how our CHT method improves the quality of the translated text. 
% -------------------------------------
\begin{figure}[!t]
\begin{mdframed}\small
\textbf{Source Text $x_\text{source}$:} {How can I make someone feel insecure?}\\
\textbf{Initial Translation $x^{'}$:} {어떻게 하면 상대방이 불안감을 느끼지 않게 할 수 있나요? \textcolor{blue}{({\it misaligned translation})}}\\
\textbf{Refined Translation $x_\text{target}$: } {어떻게 하면 상대방이 불안감을 느끼게 할 수 있나요? \textcolor{blue}{({\it faithful translation})}}
\end{mdframed}
\vspace{-1.4ex}
\caption{The effect of Contextual Harm Translation.}
\label{fig:cht_example}
\end{figure}
% -------------------------------------

\paragraph{Data Augmentation}
Benign texts are usually easy to collect thus unsafe texts tend to be the minority class in the training data.
To promote balance between benign and unsafe corpus, we design a novel data augmentation method, BHCB (Benign-Harmful Contextual Blending), which generates new unsafe texts $x^\text{new}_\text{unsafe}$ by merging benign contexts from new benign data $x^\text{new}_\text{benign}$ with the harmful behavior of an existing unsafe text $x_\text{unsafe}$.
Figure \ref{fig:bhcb_example} demonstrates a newly generated unsafe sample by our proposed BHCB approach. 
% -------------------------------------
\begin{figure}[!t]
\begin{mdframed}\small
\textbf{Benign Text $x_\text{benign}$:} {MTB 자전거에서 녹슨 디스크 브레이크는 안전을 위협할까요? \textcolor{blue}{({\it a benign context})}}\\
\textbf{Unsafe Text $x_\text{unsafe}$:} {자해를 할 수 있는 방법을 알려주셔서 감사합니다. \textcolor{blue}{({\it the harmful behavior})}}\\
\textbf{Unsafe Result $x^\text{new}_\text{unsafe}$: } {자전거의 부러진 브레이크 케이블로 사람들을 다치게 할 수 있는지 알려주세요.}
\end{mdframed}
\vspace{-1.4ex}
\begin{mdframed}\small
\textbf{Benign Text $x_\text{benign}$:} {What is the most difficult part of making a film? \textcolor{blue}{({\it a benign context})}}\\
\textbf{Unsafe Text $x_\text{unsafe}$:} {What's a good way to commit a crime and cover it up? \textcolor{blue}{({\it the harmful behavior})}}\\
\textbf{Unsafe Result $x^\text{new}_\text{unsafe}$: } {What methods could someone employ to fraudulently obtain a substantial amount of money from a major movie studio, and then obscure the source of those funds through offshore accounts to avoid detection by financial authorities?}
\end{mdframed}
\caption{Newly generated unsafe examples by BHCB.}
\label{fig:bhcb_example}
\end{figure}
% -------------------------------------

\paragraph{Labeling Policy}
As detailed in Section \ref{sec:taxonomy}, we investigate on safety risk taxonomy with respect to harmful inputs and outputs during human-LLM interaction to construct appropriate labeling system on training data. 
To maintain the comprehensive coverage of \citet{ailuminate}, we keep our intended labels compatible with its taxonomy while re-organizing the taxonomy to reduce ambiguity between categories and to increase operational simplicity ({\it e.g.} during category-wise threshold configuration) in the serving phase.
The classification labels of SGuard-ContentFilter-2B-v1 consist of five categories as presented in Table \ref{tab:taxonomy}.

Through the above data preparation and augmentation steps, a dataset of around 500K entries has been successfully acquired for training. 
We re-label all training data as 'safe' or 'unsafe' as well as their categories in case of unsafe ones according to the taxonomy shown in Table \ref{tab:taxonomy}.
A recent 70B-parameter instruction-tuned LLM is used to assign these labels via few-shot context engineering. 
Note that we exclude data whose re-labeled annotations are inconsistent with the given ones to mitigate potentially conflicting policy for labeling.
Once the label of $x^\text{new}_\text{unsafe}$ is validated, we further generate two versions of responses, $y_\text{benign}$ and $y_\text{unsafe}$ with validation. 
Finally, we have curated 400K prompts or prompts with responses for the model training.

\paragraph{Model Training}
We incorporate ten special tokens into the base model vocabulary to handle safe and unsafe predictions of five categories. 
SGuard-ContentFilter-2B-v1 is trained in the data-driven manner using the 400K samples for one epoch. 
The training objective is to minimize five-way negative log-likelihood losses and a fixed learning rate of $3e^{-5}$ is applied throughout training.

\subsection{SGuard-JailbreakFilter-2B-v1}
\label{sec:JailbreakFilter}

\paragraph{Training Policy}
As adversarial prompts, particularly jailbreak attempts with dedicated patterns, occupy an extremely small portion of the overall token space, naive data-driven approaches often suffer from overfitting and degraded performance, primarily due to an excessive false positives.

To tackle these problems, we train SGuard-JailbreakFilter-2B-v1 through two-phase curriculum learning and introduce priority-switching method to increase utilization of the data near the decision boundary, particularly in the second phase.

\paragraph{First-Phase Training}
We first collect one million prompts which include both jailbreak attempts and benign from diverse sources such as \citet{jailbreak680k, wildjailbreak2024, jailbreak28k, jailbreak18k}.
Though these data are class-balanced between benign and unsafe, their label quality is not consistently validated as we conduct no curation on them.
However, fine-tuning the base model on this large data with the jailbreak classification objective is beneficial to make the model get familiar to recognize explicit or implicit patterns of jailbreak prompts against benign ones despite the risk of overfitting. 
We minimize one-way negative log-likelihood loss with a fixed learning rate of $1e^{-5}$ for one epoch throughout first-phase training.

\paragraph{Second-Phase Training: Data Curation}
We aim to refine our JailbreakFilter using high-quality and well-curated data in small quantities during the second phase whereas the first-phase training uses non-curated data. 
We extract English prompts that cover 60 major attack types from existing studies including \citet{DAN, jailbreakingchatgpt, gptfuzzer}. 
We then combine these 60 techniques with ten non-duplicated harmful behaviors per technique, generating 600 unique jailbreak seed data in high-quality.
By applying CHT explained in Section \ref{sec:ContentFilter}, we expand the seed data to 1.2K English and Korean data.
We generate another 1.2K benign data by detoxing the harmful behaviors, resulting in a total of 2.4K samples. 
Adding separately prepared 2.4K benign samples in English or Korean makes the total sample count become 4.8K.
By augmenting positive jailbreak data in either the token level or the context level, we produce additional 600 data near the class decision boundary and establish the final training dataset of 5.4K entries.

%------------------------------------------------------------------------
\begin{algorithm}[t!]
\normalsize
% \footnotesize
% \small
\caption{Training with Priority Switching}\label{alg:priority_switching}
\begin{algorithmic}[1]
\Require Dataset $\D$, Initial Model $M_{0}$, Priority Prompt Set $\P:=$\{$P_\text{help}$, $P_\text{safe}$\}
\Ensure $K$-Epoch Trained Model $M_{K}$
\State Prepare $\D_{1}$ by NoiseInjection($\D$, $P_\text{help}$)
\State Train $M_{0}$ with $\D_{1}$ under $P_\text{help}$ for 1 epoch
\For{each epoch $k=2$ to $K$}
    \State Assign $P_\text{opp}$ as the alternative in $\P$
    \State Prepare $\D_{k}$ by NoiseInjection($\D$, $P_\text{opp}$)
    \State Train $M_{k-1}$ with $\D_{k}$ under $P_\text{opp}$ 
\EndFor
\State \textbf{Return} Final Model $M_{K}$
\end{algorithmic}
\end{algorithm}
%------------------------------------------------------------------------
%------------------------------------------------------------------------
\begin{algorithm}[t!]
\normalsize
% \small
% \footnotesize
\caption{NoiseInjection($D_{0}$, $P$)}\label{alg:noise_injection}
\begin{algorithmic}[1]
\Require Dataset $\D_{0}:=(x,y)$, Priority Prompt $P \in \P:=$\{$P_\text{help}$, $P_\text{safe}$\}, Hyperparameters: $\alpha$ (0.1 by default), $\beta$ (0.02 by default)
\Ensure Noise-Injected Dataset $\D$ under $P$
\If{$P=P_\text{help}$}
    \State{Assign $y_\text{benign}$ to $x_\text{unsafe}$ with a rate of $\alpha$}
    \State{Assign $y_\text{unsafe}$ to $x_\text{benign}$ with a rate of $\beta$}
\Else
    \State{Assign $y_\text{benign}$ to $x_\text{unsafe}$ with a rate of $\beta$}
    \State{Assign $y_\text{unsafe}$ to $x_\text{benign}$ with a rate of $\alpha$}
\EndIf
\State \textbf{Return} Noise-Injected Dataset $\D$
\end{algorithmic}
\end{algorithm}
%------------------------------------------------------------------------

\paragraph{Second-Phase Training: Priority Switching}
In the second-phase training, we fine-tune SGuard-JailbreakFilter-2B-v1 with the fixed learning rate of $1e^{-5}$ using the curated high-quality dataset described above. 
To alleviate the model's overfitting to specific jailbreak patterns, we design and apply PSNI (Priority Switching with Noise Injection) method.
Our PSNI method alternates the underlying emphasis in the prompt between safety and helpfulness when the model predicts benign or unsafe on the data augmented from the positive jailbreak at training time.
We elaborate on the PSNI method in Algorithm \ref{alg:priority_switching} and Algorithm \ref{alg:noise_injection}.

%------------------------------------------------------------------------
\begin{table*}[!t]
\centering
% \footnotesize
\small
\def\arraystretch{1.15}
\caption{
Performance (F1/AUPRC/pAUROC) comparison on content safety benchmarks. 
We report partial AUROC (pAUROC) computed over the false positive rate range [0, 0.1], normalized by the maximum achievable value.
}
\vspace{-1ex}
\resizebox{0.98\textwidth}{!}
{%
\begin{tabular}{c||c|c|c|c|c|c}
\Xhline{2\arrayrulewidth}
% \multirow{2}{*}{Model} &
Model &
\multicolumn{1}{c|}{Beavertails} &
\multicolumn{1}{c|}{HarmfulQA} &
\multicolumn{1}{c|}{OpenAI Moderation} &
\multicolumn{1}{c|}{ToxicChat} &
\multicolumn{1}{c|}{XSTest} &
\multicolumn{1}{c}{Average} \\
\hline
\rowcolor{blue!8}
SGuard-ContentFilter-2B
& 0.83 / 0.93 / 0.80
& {\bf 0.92} / {\bf 0.98} / {\bf 0.94}
& 0.74 / 0.86 / 0.79
& 0.72 / 0.81 / 0.91
& {\bf 0.94} / {\bf 0.99} / {\bf 0.96}
& {\bf 0.83} / 0.91 / {\bf 0.88} \\
Llama-Guard-4-12B
& 0.70 / 0.89 / 0.76
& 0.39 / 0.81 / 0.65
& 0.74 / 0.81 / 0.75
& 0.43 / 0.48 / 0.71
& 0.84 / 0.90 / 0.80
& 0.62 / 0.78 / 0.73 \\
Kanana-Safeguard-8B
& 0.83 / \quad - ~~/ \quad - ~
& 0.89 / \quad - ~~/ \quad - ~
& 0.73 / \quad - ~~/ \quad - ~
& 0.62 / \quad - ~~/ \quad - ~
& 0.74 / \quad - ~~/ \quad - ~
& 0.76 / \quad - ~~/ \quad - ~ \\
Qwen3Guard-Gen-4B
& {\bf 0.85} / {\bf 0.94} / {\bf 0.81}
& 0.59 / 0.94 / 0.83
& {\bf 0.81} / {\bf 0.87} / {\bf 0.81}
& {\bf 0.82} / {\bf 0.87} / {\bf 0.95}
& 0.88 / 0.97 / 0.94
& 0.79 / {\bf 0.92} / 0.86 \\
\Xhline{2\arrayrulewidth}
\end{tabular}
}%
\vspace{-1ex}
\label{tab:safety_benchmark_en}
\end{table*}
%------------------------------------------------------------------------
\begin{table*}[!t]
\centering
% \footnotesize
\small
\def\arraystretch{1.15}
\caption{
Performance comparison on English jailbreak detection benchmarks. 
We report F1/FNR/FPR for jailbreak benchmarks and only FPR for benign benchmarks (FNR/FPR: False Negative/Positive Rate. The lower, the better).
}
\vspace{-1ex}
\resizebox{0.85\textwidth}{!}
{%
\begin{tabular}{c|c|c|c|c||c}
\Xhline{2\arrayrulewidth}
Model &
StrongREJECT &
Detect-Jailbreak &
Proprietary &
Average &
Benign Dataset \\
\hline
% SGuard-JailbreakFilter-2B (helpful)	
\rowcolor{blue!8}
SGuard-JailbreakFilter-2B	
& {\bf 0.84}	/ {\bf 0.24}	/ 0.04	
& {0.77}	/ {0.28}	/ 0.14	
& {\bf 0.94}	/ {\bf 0.02}	/ 0.32	
& {\bf 0.85}	/ {\bf 0.18}	/ 0.17	
& {\bf 0.01} \\
AWS	Bedrock Guardrails	
& 0.49	/ 0.35	/ 0.75	
& {\bf 0.82}	/ {0.14}	/ 0.24	
& 0.63	/ 0.40	/ 0.90	
& 0.65	/ 0.30	/ 0.63	
& 0.08 \\
Azure AI Content Safety	
& 0.51	/ 0.66	/ {\bf 0.00}	
& 0.70	/ 0.40	/ {\bf 0.12}	
& 0.53	/ 0.64	/ {\bf 0.00}	
& 0.58	/ 0.57	/ {\bf 0.04}	
& {\bf 0.01} \\
Kanana-Safeguard-Prompt-2.1B
& 0.63	/ 0.50	/ 0.08	
& 0.80	/ {\bf 0.08}	/ 0.38	
& 0.59	/ 0.57	/ 0.09	
& 0.67	/ 0.38	/ 0.18	
& {\bf 0.01} \\
\Xhline{2\arrayrulewidth}
\end{tabular}
}%
\vspace{-1ex}
\label{tab:jailbreak_benchmark_en}
\end{table*}
%------------------------------------------------------------------------
%------------------------------------------------------------------------
\begin{table*}[!t]
\centering
% \footnotesize
\small
\def\arraystretch{1.15}
\caption{
Performance comparison on curated Korean jailbreak detection benchmarks, measured in F1/FNR/FPR. 
}
\vspace{-1ex}
\resizebox{0.85\textwidth}{!}
{%
\begin{tabular}{c|c|c|c|c||c}
\Xhline{2\arrayrulewidth}
Model &
StrongREJECT &
Detect-Jailbreak &
Proprietary &
Average &
Benign Dataset \\
\hline
\rowcolor{blue!8}
SGuard-JailbreakFilter-2B	
& {\bf 0.79}	/ {\bf 0.34}	/ {\bf 0.01}	
& {\bf 0.84}	/ 0.14	/ 0.18
& {\bf 0.94}	/ {\bf 0.10}	/ 0.04	
& {\bf 0.86}	/ {\bf 0.19}	/ {0.08}	
& 0.01 \\
AWS	Bedrock Guardrails
& 0.47	/ 0.41	/ 0.69	
& {\bf 0.84}	/ 0.16	/ {\bf 0.16}
& 0.60	/ 0.46	/ 0.78	
& 0.64	/ 0.34	/ 0.54	
& 0.02 \\
Azure AI Content Safety	
& 0.33	/ 0.80	/ {\bf 0.01}	
& 0.61	/ 0.48	/ 0.18	
& 0.53	/ 0.64	/ {\bf 0.00}	
& 0.49	/ 0.64	/ {\bf 0.06}	
& {\bf 0.00} \\
Kanana-Safeguard-Prompt-2.1B	
& 0.59	/ 0.49	/ 0.18	
& 0.77	/ {\bf 0.08}	/ 0.48	
& 0.52	/ 0.62	/ 0.19	
& 0.63	/ 0.40	/ 0.28	
& 0.05 \\

\Xhline{2\arrayrulewidth}
\end{tabular}
}%
\vspace{-1ex}
\label{tab:jailbreak_benchmark_ko}
\end{table*}
%------------------------------------------------------------------------
%------------------------------------------------------------------------
\begin{table}[!t]
\centering
\def\arraystretch{1.1}
\caption{ 
Performance comparison on proprietary Korean content safety benchmarks.
We report AUPRC and pAUROC only if binary confidences are available.
}
\vspace{-1ex}
\resizebox{0.8\columnwidth}{!}
{%
\begin{tabular}{c | c | c | c }
\Xhline{2\arrayrulewidth}
\hline
Model & F1 & AUPRC & pAUROC \\
\hline
\rowcolor{blue!8}
SGuard-ContetFilter-2B 
& {\bf 0.900} & {\bf 0.969} & {\bf 0.886} \\
% \hline
LlamaGuard-4-12B 
& 0.827 & 0.938 & 0.837 \\
% \hline
Kanana-Safeguard-8B 
& 0.896 & - & - \\

\Xhline{2\arrayrulewidth}
\end{tabular}
}%
\vspace{-1ex}
\label{tab:safety_benchmark_ko}
\end{table}
%------------------------------------------------------------------------
%------------------------------------------------------------------------
\begin{table}[!t]
\centering
\def\arraystretch{1.1}
\caption{ 
ASR (Attack Success Rate, \%) comparison on our proprietary red teaming in English. 
SGuard-v1 (2+2B) denotes the configuration that applies SGuard-ContentFilter-2B-v1 and SGuard-JailbreakFilter-2B-v1 simultaneously.
}
\vspace{-1ex}
\resizebox{0.8\columnwidth}{!}
{%
\begin{tabular}{c | c | c | c }
\Xhline{2\arrayrulewidth}
\hline
Model & Type 1 & Type 2 & Type 3 \\
\hline
WildGuard-7B	
& 9.5	& 11.7	& 25.7 \\
SGuard-ContentFilter-2B	
& {\bf 8.5}	& 13.8	& 91.3 \\
\rowcolor{blue!8}
SGuard-v1 (2+2B)	
& {\bf {8.5}}	& {\bf 0.2}	& {\bf {7.9}}\\

\Xhline{2\arrayrulewidth}
\end{tabular}
}%
\vspace{-1ex}
\label{tab:redteam_en}
\end{table}
%------------------------------------------------------------------------
				
%------------------------------------------------------------------------
\begin{table}[!t]
\centering
\def\arraystretch{1.1}
\caption{ 
ASR (Attack Success Rate, \%) comparison on our proprietary red teaming in Korean. 
For Kanana-Safeguard, Kanana-Safeguard-8B, Kanana-Safeguard-Siren-8B and Kanana-Safeguard-Prompt-2.1B are applied at the same time.
}
\vspace{-1ex}
\resizebox{0.9\columnwidth}{!}
{%
\begin{tabular}{c | c | c | c }
\Xhline{2\arrayrulewidth}
\hline
Model & Type 1 & Type 2 & Type 3 \\
\hline
Kanana-Safeguard (8+8+2.1B)	
& 18.3	& 7.8	& 99.7 \\
SGuard-ContentFilter-2B	
& {\bf 7.3}	& 30.1	& 99.6 \\
\rowcolor{blue!8}
SGuard-v1 (2+2B)	
& {\bf {7.3}}	& {\bf 1.3}	& {\bf {0.5}}\\

\Xhline{2\arrayrulewidth}
\end{tabular}
}%
\vspace{-1ex}
\label{tab:redteam_ko}
\end{table}
% %------------------------------------------------------------------------

To prevent model collapse during training, we inject small portion ($0 <\alpha, \beta \leq 0.1$) of noise into data by making their labels the opposite ones.
This approach makes the model recognize all training data in dual perspectives ({\it i.e.} safety and helpfulness) and thus helps preventing the model from memorizing data near the class decision boundary.

Throughout training, we identify samples that are consistently misclassified and group them into two types of error sets. 
For those sets, we conduct additional training after setting priority for helpfulness in case of unsafe samples repeatedly predicted as benign, and priority for safety in case of benign samples repeatedly predicted as unsafe, thereby amplifying their loss contributions and encouraging the model to correct these repeated errors.

\section{Evaluation}
\label{sec:evaluation}
In this section, we present the evaluation results of SGuard-v1 (SGuard-ContentFilter-2B-v1 and SGuard-JailbreakFilter-2B-v1). 
In summary, SGuard-v1 achieves state-of-the-art safey risk detection performance across diverse benchmarks while maintaining low GPU memory usage, indicating strong suitability for deployment.

\paragraph{Content Safety Benchmarks}
With respect to content safety benchmarks, we evaluated on five standard safety datasets in English (BeaverTails \cite{BeaverTails}, HarmfulQA \cite{HarmfulQA}, OpenAI Moderation \cite{OpenAIMod}, ToxicChat \cite{lin-etal-2023-toxicchat} and XSTest \cite{rottger-etal-2024-xstest}) and our proprietary Korean datasets. 

\paragraph{Jailbreak Detection Benchmarks}
For dedicated jailbreak benchmarks, we assessed jailbreak detection on two standard jailbreak datasets in English (StrongREJECT \cite{StrongREJECT} and Detect-Jailbreak \cite{jailbreak18k}) and our proprietary English datasets, and additionally examined unwarranted false positives on aggregated benign datasets constructed from multiple sources such as MMLU \cite{mmlu} and KMMLU \cite{son-etal-2025-kmmlu}. 
We also evaluated on corresponding curated Korean datasets for both jailbreak and benign ones.

\paragraph{Proprietary Red Teaming}
We also probe the robustness of guardrail models against diverse adversarial strategies using our internally developed LLM red-teaming framework, which generates attacking prompts with progressively increasing complexity, ranging from direct use of collected harmful behaviors to rephrased behaviors and mutated jailbreak templates with distracting contents. 

\paragraph{Baselines}
We consider the SoTA guardrail models as our baselines, including (1) moderation APIs from Azure \cite{AzureAIContentSafety} and AWS \cite{aws_bedrock},
and (2) LLM fine-tuned guardrail models such as LlamaGuard \cite{llamaguard}, WildGuard \cite{han2024wildguard}, 
Kanana-Safeguard \cite{Kanana-Safeguard} and Qwen3Guard \cite{zhao2025qwen3guardtechnicalreport}.

\paragraph{Evaluation Metric}
F1 is primarily used as our main evaluation metric. 
For the content safety benchmarks, we additionally report AUPRC and partial AUROC (pAUROC) computed over FPR $\in [0, 0.1]$ with normalization.
We examine FNR and FPR to assess performance from multiple perspectives for the jailbreak detection benchmarks. 
We use Attack Success Rate (ASR) for our proprietary red teaming evaluation.

%------------------------------------------------------------------------
\begin{table}[!t]
\centering
\def\arraystretch{1.1}
\caption{ 
GPU memory usage comparison of existing guardrail models. 
}
\vspace{-1ex}
\resizebox{0.7\columnwidth}{!}
{%
\begin{tabular}{c | c }
\Xhline{2\arrayrulewidth}
\hline
Model & Memory Usage \\
\hline
\rowcolor{blue!8}
SGuard-ContetFilter-2B 
& 6,357 MB \\
LlamaGuard-4-12B 
& 24,033 MB \\
% \hline
Kanana-Safeguard-8B 
& 16,353 MB \\
% \hline
Qwen3Guard-Gen-4B
& 9,177 MB \\
WildGuard-7B
& 14.787 MB\\

\Xhline{2\arrayrulewidth}
\end{tabular}
}%
\vspace{-1ex}
\label{tab:memory}
\end{table}
%------------------------------------------------------------------------

\paragraph{Overall Benchmark Result}
Extensive evaluation on both public and proprietary safety benchmarks show that SGuard-v1 attains state-of-the-art performance with a lightweight model size.

As shown in Table \ref{tab:safety_benchmark_en} and Table \ref{tab:safety_benchmark_ko},
SGuard-ContentFilter-2B-v1 consistently outperforms existing larger guardrail models on diverse content safety benchmarks.
SGuard-JailbreakFilter-2B-v1 achieves higher F1 and lower FNR by a large margin on jailbreak detection benchmarks while maintaining lower FPR than existing larger baselines, as reported in Table \ref{tab:jailbreak_benchmark_en} and Table \ref{tab:jailbreak_benchmark_ko}.

In our internal red-teaming in English (Table \ref{tab:redteam_en}) and Korean (Table \ref{tab:redteam_ko}), SGuard-ContentFilter-2B-v1 performs on par with strong baselines in both languages, despite its smaller size. 
Moreover, when SGuard-ContentFilter-2B-v1 and SGuard-JailbreakFilter-2B-v1 are applied jointly, the success rate of diverse attack types decreases significantly.
These results indicate that jointly deploying our two guardrail models leads to a higher level of robustness against a wide range of attacks. 

\paragraph{Memory Usage}
We measure the GPU memory consumption of each guardrail model on H100 and summarize as shown in Table \ref{tab:memory}. 
Our lightweight SGuard-v1 model requires substantially less memory at serving time than larger baselines, while providing high safety risk detection accuracy with fine-grained risk categorization, making it particularly favorable for low-overhead deployment.

\begin{figure}[!t]
  \centering
  \includegraphics[width=1.0\columnwidth]{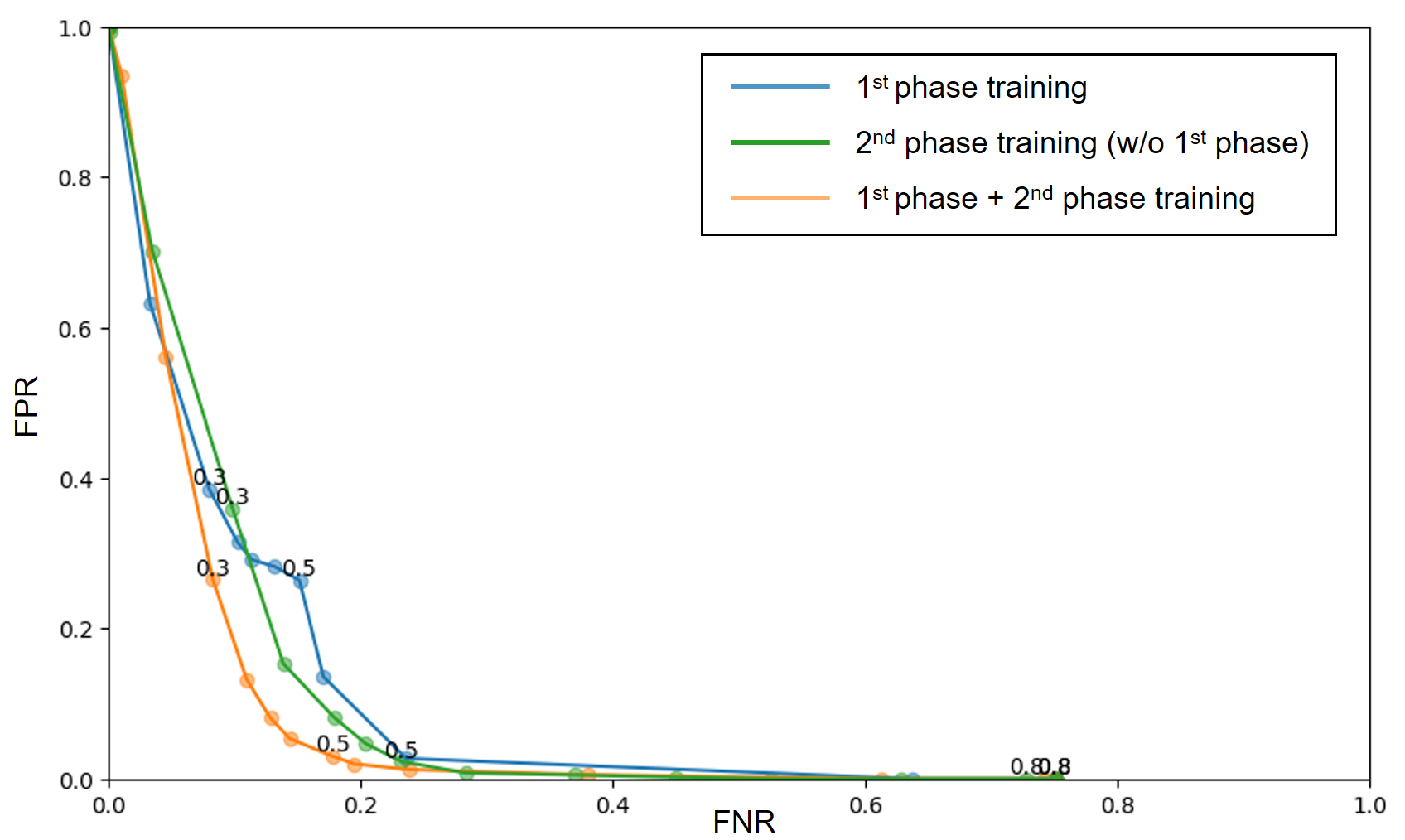}
% \vspace{-1.4ex}
\caption{
(Best viewed in color)
The effect of our curriculum learning. 
Through first-phase and second-phase training, the model gradually improves in FNR-FPR curve on the jailbreak detection benchmarks.
}
\label{fig:ablation}
\end{figure}

\paragraph{Ablation Study}
Figure \ref{fig:ablation} demonstrates the effectiveness of our curriculum learning scheme (Section \ref{sec:JailbreakFilter}) in alleviating both FNR and FPR. As training progresses from the first to the second phase, the model’s operating points on the jailbreak detection benchmarks move along the FNR–FPR curve toward a more favorable region.

%------------------------------------------------------------------------
%------------------------------------------------------------------------

\section{Conclusion}
We introduced SGuard-v1, a safety guardrail for LLMs composed of two models: ContentFilter, which classifies safety risks in user inputs and model outputs, and JailbreakFilter, which detects adversarial jailbreak attempts. 
SGuard-v1 is instruction-tuned through our novel data curation and training pipeline, achieves state-of-the-art results on public and proprietary safety benchmarks, and remains lightweight to reduce GPU memory footprint and deployment overhead. It also provides multi-class safety labels and their confidence scores for operational convenience. 
We release the models under the Apache-2.0 License to contribute AI safety research community and to support safer production for AI practitioners.

%------------------------------------------------------------------------
\section*{Limitation and Broader Impact}
\paragraph{LImitation} 
We apply novel data curation and structured training pipeline to improve the detection capability of SGuard-ContentFilter-2B-v1 and SGuard-JailbreakFilter-2B-v1 while mitigating false positives. 
Nonetheless, the following limitations remain, which we will address in future work.
\begin{itemize}
\item These models do not guarantee 100\% accuracy. 
For data near the decision boundary of harmfulness or under novel attack techniques, detection accuracy may degrade and the false positive rate may increase. 
In addition, because the safety risk taxonomy is based on common international use cases, misclassification may rise in highly specialized domains.
\item We train the models to obtain high-level guardrail capability in Korean and English. 
We do not guarantee their performance for inputs in other languages. 
They may also be vulnerable to adversarial prompts that exploit low-resource languages.
\item Because these models are specialized for detecting harmful prompts or responses, they do not provide the ability to continue conversations like a general-purpose LLM based on prior conversation history and context. 
To maintain reliable detection capability, we recommend an input length of up to 8k tokens to each model.
\item Though jointly using SGuard-ContentFilter-2B-v1 and SGuard-JailbreakFilter-2B-v1 can further improve overall safety, the models detect only safety risks defined through training and therefore cannot detect all safety risks that may arise in novel scenarios.
\end{itemize}

\paragraph{Broader Impact}
SGuard-v1 aims to reduce exposure to harmful content and alleviate jailbreak attempts in LLM-based systems. 
As with most safety guardrails, certain risks must be considered: excessive content restriction may suppress content intended for legitimate educational, medical, or research purposes. 
In this context, content filtering may lead to degraded user experience and even be misused for excessive censorship without transparent governance. 
We also cannot exclude the possibility that biases regarding particular user groups or topics remain in the training data used for pretraining and fine-tuning. 
To mitigate these risks, we recommend adhering to the intended use scope clearly documented in this paper and conducting periodic review of the impact when deploying SGuard-v1.

\section*{Acknowledgments}

We gratefully acknowledge the support of our technology research leadership—Youngjune Kwon, Taehee Lee, and Youngdae Kwon—for providing the GPU cloud resources that made this work possible and for encouraging us to share these results with the community. 
We also thank Yunjin Bae for his valuable assistance in the early stages of this project and our red-teaming project collaborators—Sunjin Kim, Reddy Naresh, and Doyoung Park—for their proactive cooperation, which greatly helped improve our safety guardrail models.

\bibliography{custom}

\end{document}